\documentclass[conference]{IEEEtran}
\IEEEoverridecommandlockouts
\usepackage{cite}
\usepackage{amsmath,amssymb,amsfonts}
\usepackage{algorithmic}
\usepackage{graphicx}
\usepackage{textcomp}
\usepackage{xcolor}
\usepackage{hyperref} 
\usepackage{textalpha} 
\usepackage{multirow}
\usepackage{booktabs}
\usepackage{siunitx}

\def\BibTeX{{\rm B\kern-.05em{\sc i\kern-.025em b}\kern-.08em
    T\kern-.1667em\lower.7ex\hbox{E}\kern-.125emX}}
\begin{document}

\title{Stress Estimation in Elderly Oncology Patients Using Visual Wearable Representations and Multi-Instance Learning\\
}

\author{
\IEEEauthorblockN{Ioannis Kyprakis}
\IEEEauthorblockA{\textit{ECE Dept. HMU}\\ 
\textit{FORTH-ICS}\\
Heraklion, Greece\\
ikyprakis@ics.forth.gr}
\and
\IEEEauthorblockN{Vasileios Skaramagkas}
\IEEEauthorblockA{\textit{ECE Dept. HMU}\\ 
\textit{FORTH-ICS}\\
Heraklion, Greece\\
vskaramag@ics.forth.gr}
\and
\IEEEauthorblockN{Georgia Karanasiou}
\IEEEauthorblockA{\textit{FORTH-BRI}\\
Ioannina, Greece\\
g.karanasiou@gmail.com}
\and
\IEEEauthorblockN{Vasilis Bouratzis}
\IEEEauthorblockA{\textit{Ioannina Univ. Hosp.,}\\
\textit{UoI Med. Sch.}\\
Ioannina, Greece\\
v.bouratzis@gmail.com}
\and
\IEEEauthorblockN{Andri Papakonstantinou}
\IEEEauthorblockA{\textit{Breast Center}\\
\textit{Karolinska Univ. Hosp.}\\
Stockholm, Sweden\\
andri.papakonstantinou@ki.se}
\and
\IEEEauthorblockN{Dimitar Stefanovski} 
\IEEEauthorblockA{\textit{Inst. of Oncology Ljubljana}\\
Ljubljana, Slovenia\\
dstefanovski@onko-i.si}
\and
\IEEEauthorblockN{Kalliopi Keramida}
\IEEEauthorblockA{\textit{NKUA Med. Sch.; Attikon}\\
\textit{Hosp., Agios Savvas Hosp.}\\
Athens, Greece\\
keramidakalliopi@hotmail.com}
\and
\IEEEauthorblockN{Aristofania Simatou} 
\IEEEauthorblockA{\textit{Agios Savvas Hosp.,}\\
Athens, Greece\\
ariasimatou@yahoo.com}
\and
\IEEEauthorblockN{Ketti Mazzocco}
\IEEEauthorblockA{\textit{IEO Psychology Division; Univ.}\\
of Milan, Italy\\
ketti.mazzocco@ieo.it}
\and
\IEEEauthorblockN{Anastasia Constantinidou}
\IEEEauthorblockA{\textit{BOCOC}\\
Nicosia, Cyprus\\
anastasia.constantinidou\\
\textit{@bococ.org.cy}
}
\and
\IEEEauthorblockN{Konstantinos Marias}
\IEEEauthorblockA{\textit{ECE Dept. HMU}\\
\textit{FORTH-ICS}\\
Heraklion, Greece\\
kmarias@ics.forth.gr}

\and
\IEEEauthorblockN{Dimitrios I. Fotiadis,}
\IEEEauthorblockA{\textit{Fellow, IEEE, FORTH-BRI}\\
Ioannina, Greece\\
fotiadis@uoi.gr}
\and
\IEEEauthorblockN{Manolis Tsiknakis}
\IEEEauthorblockA{\textit{ECE Dept. HMU}\\ 
\textit{FORTH-ICS}\\
Heraklion, Greece\\
tsiknaki@ics.forth.gr}
}

\maketitle

\begin{abstract}
Psychological stress is clinically relevant in cardio-oncology, yet it is typically assessed only through patient-reported outcome measures (PROMs) and is rarely integrated into continuous cardiotoxicity surveillance. We estimate perceived stress in an elderly, multi-center breast cancer cohort (CARDIOCARE) using multimodal wearable data from a smartwatch (physical activity and sleep) and a chest-worn ECG sensor. Wearable streams are transformed into heterogeneous visual representations, yielding a weakly supervised setting in which a single Perceived Stress Scale (PSS) score corresponds to many unlabeled windows. A lightweight pretrained mixture-of-experts backbone (Tiny-BioMoE) embeds each representation into 192-dimensional vectors, which are aggregated via attention-based multi-instance learning (MIL) to predict PSS at month 3 (M3) and month 6 (M6). Under leave-one-subject-out (LOSO) evaluation, predictions showed moderate agreement with questionnaire scores (M3: $R^2{=}0.24$, Pearson $r{=}0.42$, Spearman $\rho{=}0.48$; M6: $R^2{=}0.28$, Pearson $r{=}0.49$, Spearman $\rho{=}0.52$), with global RMSE/MAE of 6.62/6.07 at M3 and 6.13/5.54 at M6.
\newline

\textbf{Clinical Relevance—}
\textbf{This work provides a scalable pathway to estimate patient-reported stress from passively collected wearable data in elderly cardio-oncology, enabling longitudinal monitoring between clinic visits. Even moderate predictive skill can support holistic surveillance by contextualizing stress alongside sleep disruption, activity decline, and cardiac dynamics during cardiotoxic treatment cycles.}
\end{abstract}


\section{Introduction}

Psychological stress is a significant concern for elderly individuals, particularly those facing cancer or cardiovascular disease. Chronic stress disrupts the regulation of nervous, immune, and endocrine systems, leading to increase in disease risk, decrease quality of life (QoL), and worsening of health outcomes \cite{McEwen1998, Steptoe2012}. In cancer patients, higher stress levels are associated with lower adherence to treatment, greater difficulty in managing side effects, and faster decline in physical function \cite{Paslaru2025}. These burdens are even more pronounced in older adults, who often have multiple health conditions and reduced physical resilience.

Cardio-oncology care prioritizes identification of risks and monitoring of heart complications caused by cancer treatments, in line with current guidelines recommending assessments before, during, and after therapy \cite{Lyon2022}. Studies of cardiotoxicity in older breast cancer patients show that these individuals face unique challenges and risk factors, often requiring close surveillance—especially for those experiencing experiencing  significant comorbidities, cardiovascular risk factors, established cardiovascular disease, frailty, other illnesses, or reduced independence \cite{Keramida}. Because elderly breast cancer patients are particularly susceptible to cardiovascular events and loss of functionality, broader monitoring strategies are needed to support care tailored to their needs.

In cardio-oncology, psychological stress is increasingly recognized as a modifier of cardiovascular vulnerability, treatment tolerance, and recovery, interacting with autonomic dysregulation, sleep disturbance, and physical deconditioning during cardiotoxic therapy~\cite{Rozanski, Tsuji}. Stress also affects crucial health behaviors, such as sleep, physical activity, and adherence to prescribed therapies, which are vital during cancer treatments that may harm the heart \cite{Antoni2023}. Yet, stress remains under-monitored in routine cardio-oncology care, mostly because assessments rely on infrequent self-reports and sporadic clinic visits which do not capture day-to-day changes in stress levels \cite{Pinge}.

Current advancements in wearable technology now enable ongoing, real-world monitoring of stress-related physiological signals, including heart rate variability, activity, and sleep patterns \cite{Smets2018, Dunn2018}. These data streams make it possible to move beyond basic summaries and develop digital biomarkers more closely reflecting how the body manages stress. However, traditional analysis methods often struggle to capture the complex, dynamic, and context-specific nature of stress responses, especially in older patients with diverse health backgrounds and sometimes incomplete data \cite{Yang2025}.

Recent work converts physiological time series into images (e.g., spectrograms, recurrence plots) to leverage vision backbones \cite{Gkikas2025MultiRep}. Vision-based learning approaches have shown strong performance in affective computing tasks such as stress, fatigue, and sleep assessment by capturing temporal dynamics and multi-scale patterns which are challenging to encode manually \cite{Ziaratnia2024CCTLSTM}. However, most existing studies concentrate on healthy cohorts, utilize single-modality data, and neglect practical constraints relevant to clinical deployment, such as model size, interpretability, and robustness \cite{Vos}.

A further fundamental challenge in modeling stress using wearable data is its weakly supervised nature. Stress labels are primarily available at sparse temporal resolutions, often derived from questionnaires or periodic assessments, whereas physiological signals are continuously recorded \cite{Pinge}. Multi-instance learning (MIL) provides a well-suited approach for addressing this discrepancy by enabling bag-level predictions from collections of unlabeled instances and has been demostrating promising potential in medical imaging and biosignal analysis \cite{Ilse2018MIL}. Nevertheless, the application of MIL to visual representations of wearable data in elderly clinical populations remains insufficiently explored.

Concurrently, the demand for lightweight and efficient models is increasingly recognized in digital health, where computational resources, energy efficiency, and scalability are essential for clinical implementation. Mixture-of-Experts (MoE) architectures offer modularity and specialization but are frequently computationally intensive \cite{Huang}. Recent advances in compact and resource-efficient expert models indicate that well-designed lightweight MoE variants can maintain robust performance while remaining suitable for deployment \cite{Fedus2022}. However, these approaches have seldom been examined in biomedical stress modeling, especially in conjunction with MIL and visual representations.

Motivated by this gap, this study explores stress prediction in an elderly cardio-oncology cohort as part of the EU-funded clinical study CARDIOCARE~\cite{CARDIOCARE}, using time-series data from wearable devices and advanced deep learning techniques. We focus on perceived stress assessed via PROMs at sparse follow-up timepoints, while wearable signals are continuously recorded in free-living conditions.

To address this weak-supervision regime, we propose a novel multimodal framework that combines BioMoE architectures with attention-based MIL to infer stress from diverse wearable streams when labels are limited. The approach leverages complementary wearable-derived representations and learns to aggregate variable-length, partially missing evidence into patient-level predictions, while providing instance-level interpretability through attention. By targeting a clinically meaningful and under-studied cardio-oncology population, this work advances scalable, non-invasive tools for personalized monitoring, early risk identification, and quality-of-life support in older patients undergoing cancer treatment.

\section{Related Work}

In cardio-oncology and cardiotoxicity-related cohorts, stress estimation models remain notably underexplored. Clinical work prioritizes cardiovascular endpoints, while psychosocial stress is typically captured only through self-reported questionnaires. Recent reviews emphasize both the vulnerability of elderly breast cancer patients and the lack of continuously deployable, quantitative stress-estimation tools embedded in cardio-oncology monitoring pathways \cite{Nechita2025}. To the best of our knowledge, no studies report predictive performance for stress estimation models developed or validated specifically in cardio-oncology populations.

Outside oncology, stress estimation commonly uses handcrafted features (time/frequency/statistical) from wearable streams, followed by conventional classifiers, with heart rate variability (HRV) and electrodermal activity (EDA) repeatedly identified as highly informative. In controlled settings, near-ceiling performance is reported: on WESAD (15 healthy subjects; ECG, EDA, respiration, accelerometry), Calvo et al. \cite{Calvo} achieved 99.8\% accuracy with CNNs (personalized: 99.6\%), attributing major contribution to HRV and EDA; Heyat et al. \cite{Heyat2022} reported 93.3\% accuracy (F1 93.5\%) intra-subject and 94.1\% inter-subject using ECG-derived HRV features from a smart t-shirt study (20 participants). Multimodal fusion further improves robustness: Amin et al. \cite{Amin2025} showed HRV+EDA consistently outperformed HRV-only across devices (35 participants), with AUROC up to 0.961 for a consumer smartwatch and performance approaching 0.98 when EDA was included.

In contrast, ecological validity remains limited. Martinez et al. \cite{Martinez2022} monitored 657 information workers for eight weeks and found that HRV features explained only ~1\% of variance in momentary self-reported stress (max ~2.2\% with aggregation), underscoring the gap between laboratory accuracy and in-the-wild stress estimation. Recent visual-encoding approaches map physiological time series to images to leverage vision models: Arya et al. \cite{Arya2025} used fuzzy recurrence plots from HRV and achieved 96.6\% accuracy with an SVM classifying relaxation states, while Yang et al. \cite{Yang2025} used Gramian/Markov fields on WESAD and reported 90.96\% accuracy and 91.67\% F1. Despite promising results, such visual pipelines remain largely unexplored in elderly cardio-oncology cohorts, where labels are weakly supervised (questionnaire-level), and continuous wearable streams require principled aggregation beyond heuristic windowing

\section{Methodology}

\subsection{Dataset}

The proposed framework was evaluated on data from 387 patients, recruited within the CARDIOCARE clinacal study. The cohort consists of elderly women, over sixty years of age, diagnosed with breast cancer and considered at risk of cancer therapy–related cardiotoxicity. Data were collected across six clinical centers: Bank of Cyprus Oncology Center (BOCOC, No. ΕΕΒΚ/ΕΠ/2022/58), Karolinska Universitetssjukhuset (KSBC, No. 2023-0062-01-413152),  University of Ioannina (UOI, No. 25/23-11/2022), National and Kapodistrian University of Athens (NKUA, No. 456/14-10-2022, ΕΒΔ-683/22-11-2022), Onkoloski Institut Ljubljana (IOL, No. ERIDEK-0038/2023), and Instituto Europeo Di Oncologia SRL (IEO, No. R1754/22-IEO 1874). Written informed consent was obtained from all participants prior to enrolment and approvals were received from the relevant Ethics Committees at all participating clinical centers.

Multimodal data were acquired using wearable and clinical-grade sensors. Physical activity and sleep data were collected via the Garmin Venu SQ smartwatch \cite{GarminVenuSQ}, with a minimum availability of at least two days per week per participant. In parallel, electrocardiographic (ECG) data were collected using the Polar H10 (Fs = 130Hz) chest strap \cite{PolarH10}, consisting of at least 30-minute recordings acquired biweekly during the first six months following enrolment.

Stress assessment was performed using patient-reported outcome measures (PROMs), with perceived stress quantified using the Perceived Stress Scale (PSS) \cite{Cohen1983}. The PSS was completed at baseline (M0), month 3 (M3), and month 6 (M6). In this study, PSS scores collected at M3 and M6 were used as continuous regression targets for stress estimation, representing participants’ self-reported perceived stress at each assessment timepoint.
This acquisition protocol and follow-up schedule were designed by the project’s expert oncologists to align measurements with the project’s clinical monitoring requirements.

\subsection{Preprocessing}

Multimodal wearable data from smartwatches and chest-worn ECG sensors were preprocessed and transformed into visual representations suitable for vision-based learning. Given the real-world, longitudinal nature of the data and the elderly clinical cohort, preprocessing focused on temporal structuring, handling of aggregation-level missingness, and standardized visual encoding, rather than on aggressive signal-level denoising.

Physical activity and sleep data were processed independently but followed a common weekly aggregation strategy. For each patient, the earliest available calendar date was used as the temporal baseline, and all subsequent smartwatch recordings were aligned to this reference. Daily summaries were grouped into non-overlapping weekly segments, and weeks with more than 60\% missing daily values were excluded from further analysis. For retained weeks, missing daily entries were imputed using feature-wise means within the same week. All features were then z-score normalized across the seven days of each week, emphasizing relative intra-week temporal patterns rather than absolute magnitudes.

\begin{figure*}[!htbp]
\centering
\includegraphics[width=\linewidth]{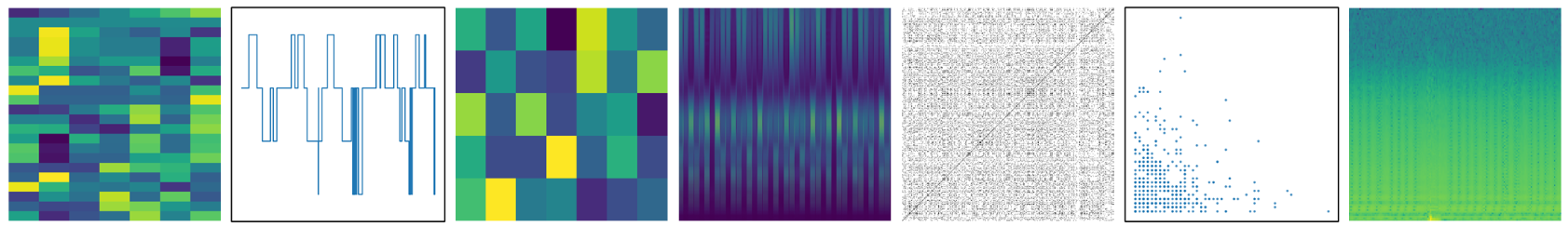}
\caption{Example of visual representations used. From left to right: Physical Activity Heatmap, Hypnogram, Sleep Heatmap, Scalogram, Recurence, Poincaré, Spectrogam}
\label{figure1}
\end{figure*}

For physical activity, daily summary features were arranged into a feature-by-day matrix for each patient-week and converted into minimal heatmap images (Fig.~\ref{figure1}), with rows corresponding to activity-related features and columns to days of the week. Axes, ticks, and annotations were removed to ensure that the models focus exclusively on structural patterns.

Sleep was represented using two complementary visual encodings. First, weekly sleep feature heatmaps were generated from nightly sleep summaries, using sleep duration, unmeasurable sleep duration, deep sleep duration, light sleep duration, and REM sleep duration as features. For each patient and week, these nightly values were mapped to the corresponding day of the week, forming a compact feature-by-day matrix. Feature-wise normalization across the week was applied, and the resulting matrices were rendered as minimal, axis-free heatmaps. Second, sleep hypnogram images were generated from epoch-level sleep stage annotations, capturing intra-night sleep architecture and stage transitions as compact step-function images without axes or labels.

ECG signals were segmented into 5-minute non-overlapping windows, and windows shorter than 5 minutes were excluded. For each ECG window, signal quality was first assessed using the ECG quality assessment function provided by the NeuroKit \cite{neurokit} toolkit. Subsequently, multiple complementary visual representations were generated (Fig.~\ref{figure1}), including recurrence plots, spectrograms, scalograms, and Poincaré plots, to capture nonlinear temporal dependencies, time–frequency characteristics, multi-scale oscillatory patterns, and beat-to-beat variability dynamics. All ECG visualizations were rendered in a minimal format without axes or annotations to ensure consistency across modalities.

All smartwatch and ECG derived visual instances were assigned to the M3 or M6 PSS assessment according to their time from the patient-specific baseline. To prevent temporal leakage, this alignment was not performed symmetrically: instances recorded after the month 3 assessment were never assigned the month 3 label, and month 3 labels were derived only from data available up to the M3 visit. Since multiple visual instances correspond to a single questionnaire-derived stress score, this setup naturally results in a weakly supervised learning problem, in which bag-level stress labels are associated with sets of unlabeled visual instances. This formulation motivated the use of multi-instance learning in the subsequent modeling pipeline. Table~\ref{tab:visual_matrix_stress} summarizes the distribution of the final dataset used for stress estimation at M3 and M6, detailing the number of patient-level bags and corresponding visual representation instances per modality and stress level. 

\begin{table}[htbp]
\caption{Distribution of visual representation instances per modality, stress level, and time point.}
\label{tab:visual_matrix_stress}
\centering
\renewcommand{\arraystretch}{1.2}
\begin{tabular}{|l|l|c|c|c|c|}
\hline
\textbf{Modality} & \textbf{Representation} &
\multicolumn{2}{c|}{\textbf{M3}} &
\multicolumn{2}{c|}{\textbf{M6}} \\
\cline{3-6}
 &  & \textbf{Low} & \textbf{Elev.} & \textbf{Low} & \textbf{Elev.} \\
\hline
Activity & Weekly heatmaps & 2028 & 2721 & 1629 & 1943 \\
\hline
\multirow{2}{*}{Sleep}
 & Weekly heatmaps & 1162 & 2071 & 1067 & 1344 \\
 & Hypnograms & 9464 & 11532 & 7647 & 9298 \\
\hline
\multirow{4}{*}{ECG}
 & Recurrence plots
 & \multirow{4}{*}{3102}
 & \multirow{4}{*}{5310}
 & \multirow{4}{*}{1873}
 & \multirow{4}{*}{2715} \\
 & Spectrograms &  &  &  &  \\
 & Scalograms &  &  &  &  \\
 & Poincar\'e plots &  &  &  &  \\
\hline
\end{tabular}
\end{table}

\subsection{Model Architecture and Instance-Level Processing}

\begin{figure*}[!htbp]
\centering
\includegraphics[width=\linewidth]{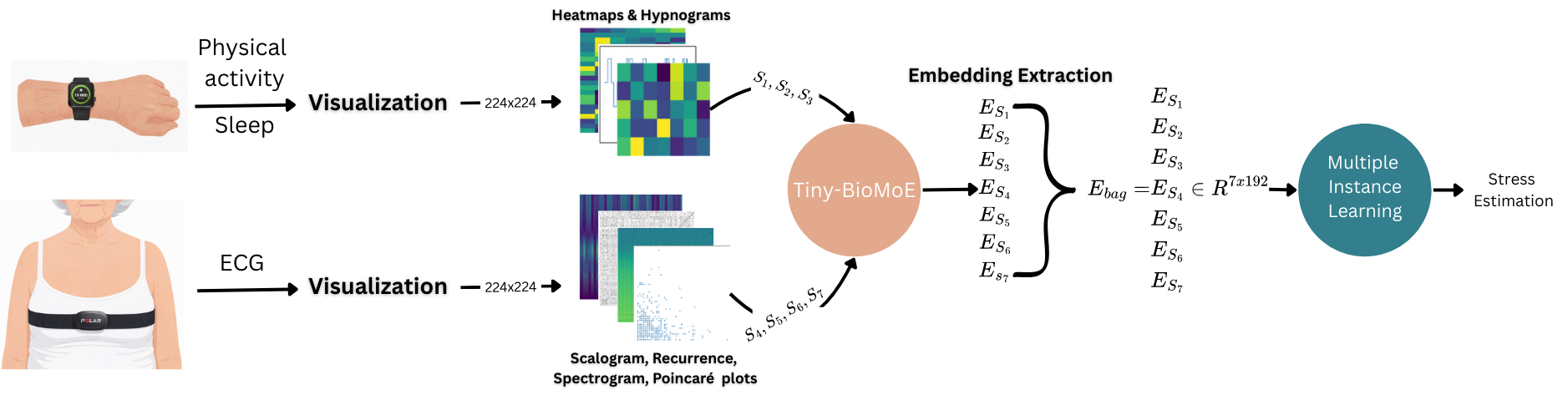}
\caption{A schematic illustration of the proposed pipeline; The first two images (human icons) were generated using Google Gemini \cite{Gemini} (synthetic images).}
\label{figure2}
\end{figure*}

Our framework follows a weakly supervised MIL setup aligned with Fig.~\ref{figure2}. For each patient $p$ and follow-up horizon $h \in \{\mathrm{M3}, \mathrm{M6}\}$, we construct a bag of $N_{p,h}$ visual instances extracted from wearable data (weekly smartwatch images and windowed ECG images). Each instance is a $224\times224$ RGB image and is associated with a modality identifier $m_i \in \{0,1,2\}$ corresponding to ECG, physical activity, and sleep, respectively.

\paragraph{Tiny-BioMoE embedding extraction}
Each instance image $\mathbf{x}_i \in \mathbb{R}^{3\times224\times224}$ is mapped to a compact representation using a pretrained lightweight mixture-of-experts backbone (Tiny-BioMoE) \cite{tiny}. The backbone comprises two complementary encoders: a SpectFormer-based encoder $f_{1}(\cdot)$ and an EfficientViT-based encoder $f_{2}(\cdot)$, each producing a $d$-dimensional vector with $d=96$. Following input-level normalization, each encoder output is (i) layer-normalized, and (ii) modulated by a lightweight gating function implemented as a small feed-forward module. The final instance embedding is obtained by concatenation and output layer normalization:
\begin{equation}
\begin{split}
\mathbf{e}_i
=\mathrm{LN}_{\text{out}}\!\Big(
\big[\tilde{\mathbf{z}}_{i}^{(1)};\tilde{\mathbf{z}}_{i}^{(2)}\big]
\Big),\\
\tilde{\mathbf{z}}_{i}^{(k)}
=\mathrm{LN}_{k}\!\left(\mathbf{z}_{i}^{(k)}\right)\odot
g\!\left(\mathbf{z}_{i}^{(k)}\right),\\
\mathbf{z}_{i}^{(k)}
=f_k\!\left(\mathrm{LN}_{\text{img}}(\mathbf{x}_i)\right).
\end{split}
\label{eq:tinybiomoe}
\end{equation}
where $\mathbf{e}_i\in\mathbb{R}^{192}$ is the final instance-level embedding, $[\cdot;\cdot]$ denotes concatenation, $\mathrm{LN}(\cdot)$ is layer normalization, and $\odot$ denotes element-wise multiplication.
The gating function $g(\cdot)$ corresponds to the lightweight module used in our implementation (ELU $\rightarrow$ linear $\rightarrow$ $\mathrm{Hardtanh}(0,1)$), producing a bounded, per-dimension modulation vector.
In Phase~1, for each patient--horizon bag $(p,h)$ we store the set of embeddings $\{\mathbf{e}_i\}_{i=1}^{N_{p,h}}$ together with modality identifiers $\{m_i\}_{i=1}^{N_{p,h}}$ (0=ECG, 1=physical activity, 2=sleep). We extract instances from all available view types per modality (ECG: recurrence/spectrogram/scalogram/Poincar\'e; physical activity: weekly maps; sleep: hypnogram and weekly views), and save one compressed file per bag containing \texttt{embeddings} and \texttt{modality\_ids}.

\paragraph{Attention-based MIL aggregation}
Given a bag of instance embeddings $\{\mathbf{e}_i\}_{i=1}^{N_{p,h}}$, we compute a patient--horizon representation using attention-based MIL. Each instance embedding is first-layer-normalized and augmented with a learned modality embedding:
\begin{equation}
\tilde{\mathbf{e}}_i = \mathrm{LN}(\mathbf{e}_i) + \mathbf{v}_{m_i},\qquad
\mathbf{h}_i=\phi(\tilde{\mathbf{e}}_i)\in\mathbb{R}^{D},
\label{eq:modemb}
\end{equation}
where $\mathbf{v}_{m}\in\mathbb{R}^{192}$ is implemented as an embedding table of size $3\times192$. We then compute an attention logit per instance and normalize attention weights within each bag:
\begin{equation}
\ell_i = a(\mathbf{h}_i), \qquad
\alpha_i=\frac{\exp(\ell_i)}{\sum_{j=1}^{N_{p,h}}\exp(\ell_j)}.
\label{eq:attn}
\end{equation}
The bag representation is the attention-weighted sum,
\begin{equation}
\mathbf{h}_{p,h}=\sum_{i=1}^{N_{p,h}} \alpha_i \mathbf{h}_i,
\label{eq:bagrepr}
\end{equation}
and the regression head outputs the stress estimate:
\begin{equation}
\hat{y}_{p,h}=\psi(\mathbf{h}_{p,h}).
\label{eq:pred}
\end{equation}
\noindent\textbf{Architecture details.}
We set the instance latent dimension to $D=256$. The instance projector $\phi(\cdot)$ is a two-layer MLP that maps the 192-d embedding into a fixed 256-d instance space ($192 \rightarrow 256$; ELU + dropout $p=0.15$), enabling attention pooling across variable-length bags. The attention scorer $a(\cdot)$ is a lightweight MLP ($256 \rightarrow 128 \rightarrow 1$) with $\tanh$, producing one unnormalized attention logit per instance. Attention weights are obtained via a bag-wise (segment-wise) softmax over instance-to-bag indices, so normalization is performed independently within each bag and naturally supports variable-length multimodal instance sets. Finally, the regression head $\psi(\cdot)$ maps the pooled bag representation through a two-layer MLP ($256 \rightarrow 128 \rightarrow 1$; ELU + dropout $p=0.15$) to predict the continuous PSS score.

\section{Experiments and Results}

\subsection{Evaluation Protocol}

Following our evaluation design, we adopt a leave-one-subject-out (LOSO) split at the patient level to prevent information leakage across horizons and visual representations. In each LOSO iteration, one patient is held out for testing, while the remaining patients form the training pool and are deterministically partitioned into training and validation subsets (80\%--20\%) at the patient level. Early stopping monitors validation RMSE on this held-out validation subset, and the best-performing checkpoint (minimum validation RMSE) is used to evaluate the test patient. We run two horizon settings (M3$\rightarrow$M3 and ALL$\rightarrow$M6) and report modality ablations, including ALL, PHYS+SLEEP, PHYS+ECG, and SLEEP+ECG. All splits, initializations, and optimization procedures are performed under a deterministic setup to ensure reproducibility.

\subsection{Experimental Setup}
All experiments were implemented in PyTorch and conducted on a single GPU. Optimization was performed using the AdamW optimizer with an initial learning rate of $5\times10^{-4}$ and a weight decay of $10^{-4}$. Models were trained for a maximum of 150 epochs, with early stopping based on validation RMSE and a patience of 15 epochs.

A warm-up and cosine annealing learning rate schedule was employed, consisting of a linear warm-up phase over the first 10 epochs followed by cosine decay. Training was performed at the bag level with a batch size of 8 bags. To control memory usage and ensure consistent instance-level processing across patients, the maximum number of instances per bag was capped at 512. All hyperparameters were fixed across experiments and ablations. Model size and computational cost are summarized in Table~\ref{tab:complexity}.

\begin{table}[!htbp]
\centering
\caption{Model complexity (parameters and FLOPs).}
\label{tab:complexity}
\small
\setlength{\tabcolsep}{10pt}
\renewcommand{\arraystretch}{1.15}
\begin{tabular}{l
                S[table-format=1.3]
                S[table-format=1.2]}
\toprule
\textbf{Component} & {\textbf{Params (M)}} & {\textbf{FLOPs (G)}} \\
\midrule
Tiny-BioMoE          & 7.340 & 3.04 \\
Attention-MIL   & 0.215 & 1.21 \\
\midrule
\textbf{Total}       & \textbf{7.555} & \textbf{4.25} \\
\bottomrule
\end{tabular}
\end{table}

\subsection{Evaluation Metrics}
Stress estimation performance was primarily evaluated using root mean squared error (RMSE), reflecting the continuous nature of perceived stress scores derived from the PSS questionnaire. Mean absolute error (MAE) was additionally reported as a complementary scale-preserving global metric. For LOSO evaluation, RMSE is summarized as the mean and standard deviation across subject-level folds. In addition, global performance was computed by pooling predictions from all LOSO test folds and reporting overall RMSE, MAE, coefficient of determination ($R^2$), and linear and rank-based associations using Pearson and Spearman correlation coefficients, respectively. Global metrics are computed by pooling all out-of-subject predictions across LOSO folds, thus weighting subjects by the number of test instances. This design allows us to capture typical per-patient generalization (fold-level statistics) while also quantifying sample-weighted performance over all out-of-subject predictions (global metrics).

\subsection{Results}
The results of the proposed framework are summarized in Table~\ref{tab:results_all_m3m6}. The multimodal configuration, integrating ECG, physical activity, and sleep representations, consistently achieved the best performance across all evaluation metrics. In particular, multimodal fusion resulted in lower RMSE and MAE and higher correlation with ground-truth PSS scores compared to single-modality baselines.

\begin{table}[!t]
\caption{LOSO performance using all modalities. Global metrics are computed over pooled out-of-subject predictions.}
\label{tab:results_all_m3m6}
\centering
\small
\renewcommand{\arraystretch}{1.15}
\setlength{\tabcolsep}{7pt}
\begin{tabular}{|l|c|c|}
\hline
\textbf{Metric} & \textbf{M3} & \textbf{M6} \\
\hline
RMSE (mean$\pm$std) & 4.17$\pm$3.24 & 3.34$\pm$3.11 \\ \hline
Global RMSE & 6.62 & 6.13 \\ \hline
Global MAE  & 6.07 & 5.54 \\ \hline
Global $R^2$ & 0.24 & 0.28 \\ \hline
Global Pearson $r$ & 0.42 & 0.49 \\ \hline
Global Spearman $\rho$ & 0.48 & 0.52 \\ \hline
\end{tabular}
\end{table}

Fold-to-fold variability was moderate, suggesting reasonably consistent out-of-subject performance. Models relying solely on ECG representations outperformed smartwatch-only configurations, highlighting the importance of cardiac dynamics for stress estimation. However, the inclusion of physical activity and sleep information provided additional gains, supporting the complementary role of behavioral and physiological signals in modeling perceived stress.

\subsection{Ablation Study}
To further analyze the contribution of each modality and its interactions, we conducted a series of pairwise ablation experiments. Specifically, we evaluated models trained using combinations of (i) ECG and sleep, (ii) ECG and physical activity, and (iii) physical activity and sleep representations. Results are reported in Table~\ref{tab:ablation_results}.

\begin{table}[!htbp]
\caption{Ablation study (pairwise modality fusion) under LOSO, transposed for compactness. P=Physical activity, S=Sleep, E=ECG.}
\label{tab:ablation_results}
\centering
\small
\renewcommand{\arraystretch}{1.15}
\setlength{\tabcolsep}{7pt}
\begin{tabular}{|l|c|c|c|}
\hline
\textbf{Metric} & \textbf{P+S} & \textbf{P+E} & \textbf{S+E} \\
\hline
M3 RMSE (m$\pm$s)   & 6.00$\pm$4.98 & 6.04$\pm$4.95 & 6.10$\pm$4.94 \\
M3 Global RMSE      & 7.79          & 7.81          & 7.85 \\
\hline
M6 RMSE (m$\pm$s)   & 5.60$\pm$4.80 & 5.53$\pm$4.71 & 5.50$\pm$4.72 \\
M6 Global RMSE      & 7.09          & 7.12          & 7.15 \\
\hline
\end{tabular}
\end{table}

The pairwise ablation analysis shows modest but consistent differences across modalities. At M3, the three two-modality configurations perform similarly (RMSE $\approx$ 6.0), with physical activity-sleep slightly lower than ECG-containing pairs, while global RMSE remains comparable across all combinations. At M6, performance improves overall (RMSE $\approx$ 5.5--5.6), and the best pairwise result is obtained by sleep-ECG, followed by physical activity-ECG, with physical activity-sleep slightly worse. This pattern suggests that, at follow-up, ECG provides complementary information beyond smartwatch-derived signals, and that sleep features combine more effectively with ECG than physical activity alone; however, the margins are small and should be interpreted as incremental gains rather than categorical superiority.

\section{Discussion}

In this study, we evaluated stress estimation in a real-world cardio-oncology setting where supervision is temporally coarse: PSS is collected only at M3 and M6, while wearables provide dense multimodal signals that we convert into visual representations. This creates a weakly supervised problem in which many short sensor windows map to a single questionnaire score; results must therefore be interpreted in light of the mismatch between window-level physiology/behavior and weeks-long PSS recall.

In general, the pipeline shows moderate predictive association with questionnaire scores (M3: $R^2{=}0.24$, $r{=}0.42$, $\rho{=}0.48$; M6: $R^2{=}0.28$, $r{=}0.49$, $\rho{=}0.52$), although absolute error remains non-trivial (global RMSE/MAE 6.62/6.07 at M3 and 6.13/5.54 at M6). The gap between fold-level summaries and pooled metrics is expected because pooled evaluation weights subjects by the number of available bags and is more sensitive to adherence imbalance and difficult cases. This performance profile is consistent with stress modeling in free-living settings, where contextual and individual factors not captured by sensors can dominate perceived stress, and temporal misalignment between PSS recall and nearby wearable data increases label noise.

MIL matches the supervision structure: labels exist at the patient--horizon level, while inputs comprise many unlabeled instances per horizon. Attention-based aggregation offers a principled alternative to uniform pooling, naturally handles variable instance counts, and is compatible with standard strategies for handling missing modalities. The higher Spearman $\rho$ relative to Pearson $r$ suggests the model may capture ordinal differences (higher vs.\ lower stress) more reliably than exact PSS values, which is plausible given inter-subject variability and the coarse target.

From an oncology and cardio-oncology perspective, continuously captured stress patterns from wearable data can complement patient-reported measures by revealing early psychosocial and behavioral strain, potentially flagging patients who may later experience treatment intolerance, symptom worsening, cardiotoxicity or decline in functional capacity.

Performance is likely bounded by several factors. The cohort consists of elderly patients with comorbidities, substantial multimodal missingness and variable adherence, and multi-center effects. PSS is subjective and temporally coarse relative to the instance-generation process, and strict LOSO evaluation, while clinically appropriate, limits the benefit of patient-specific calibration.

Future work will prioritize reducing label--instance misalignment through more time-local targets (e.g., symptom diaries) and improving robustness via explicit modeling of missingness and site effects with calibrated uncertainty. Evaluation should shift from point prediction toward clinically aligned objectives, including within-subject change detection and patient-level stratification, with validation against downstream outcomes (e.g., sleep disruption, symptom burden, and treatment tolerance).

\section{Conclusion}

In this work, we investigated perceived-stress estimation in an elderly, multi-center cardio-oncology cohort using passively collected smartwatch and ECG data encoded as visual representations. Under leakage-free LOSO evaluation, the proposed multimodal approach achieved moderate agreement with PSS at both M3 and M6, indicating that vision-based wearable signatures can recover clinically relevant stress-related patterns even in heterogeneous free-living conditions. Future work will prioritize more time-local supervision and improved robustness to missingness and site effects, with evaluation centered on within-patient change detection and clinically actionable monitoring.

\vspace{12pt}

\begin{thebibliography}{00}
\bibitem{McEwen1998} B. S. McEwen, “Protective and damaging effects of stress mediators,” New England Journal of Medicine, vol. 338, no. 3, pp. 171–179, 1998, doi: 10.1056/NEJM199801153380307.

\bibitem{Steptoe2012} A. Steptoe and M. Kivimäki, “Stress and cardiovascular disease,” Nature Reviews Cardiology, vol. 9, no. 6, pp. 360–370, 2012, doi: 10.1038/nrcardio.2012.45.

\bibitem{Paslaru2025}
A. M. Paslaru \emph{et al.}, “Mind over malignancy: A systematic review and meta-analysis of psychological distress, coping, and therapeutic interventions in oncology,” \emph{Medicina}, vol. 61, no. 6, p. 1086, 2025, doi: 10.3390/medicina61061086.

\bibitem{Lyon2022}
A. R. Lyon \emph{et al.}, “Baseline cardiovascular risk assessment in cancer patients scheduled to receive cardiotoxic cancer therapies: A position statement and new risk assessment tools from the Cardio-Oncology Study Group of the Heart Failure Association of the European Society of Cardiology in collaboration with the International Cardio-Oncology Society,” \emph{European Journal of Heart Failure}, vol. 22, no. 11, pp. 1945--1960, 2020, doi: 10.1002/ejhf.1920.


\bibitem{Keramida} K. Keramida et al., “Cardiotoxicity in elderly breast cancer patients,” Cancers, vol. 17, no. 13, p. 2198, 2025, doi: 10.3390/cancers17132198.

\bibitem{Rozanski} Rozanski, A., Blumenthal, J. A., and Kaplan, J. (1999). Impact of psychological factors on the pathogenesis of cardiovascular disease. Circulation, 99(16), 2192–2217. doi: 10.1161/01.CIR.99.16.2192

\bibitem{Tsuji}  Tsuji, H. \emph{et al.}, (1996). Reduced heart rate variability and mortality risk. Circulation, 94(11), 2850–2855. doi: 10.1161/01.CIR.94.11.2850

\bibitem{Antoni2023}
M.~H. Antoni \emph{et al.}, “Stress management interventions to facilitate psychological and physiological adaptation and optimal health outcomes in cancer patients and survivors,” \textit{Annual Review of Psychology}, vol.~74, pp.~423--455, 2023, doi:~10.1146/annurev-psych-030122-124119.


\bibitem{Pinge}
A. Pinge, V. Gad, D. Jaisighani, S. Ghosh, and S. Sen, “Detection and monitoring of stress using wearables: A systematic review,” \emph{Frontiers in Computer Science}, vol. 6, Art. no. 1478851, 2024, doi: 10.3389/fcomp.2024.1478851.

\bibitem{Smets2018} E. Smets et al., “Large-scale wearable data reveal digital phenotypes for daily-life stress detection,” npj Digital Medicine, vol. 1, no. 1, p. 67, 2018, doi: 10.1038/s41746-018-0074-9.

\bibitem{Dunn2018} J. Dunn, R. Runge, and M. Snyder, “Wearables and the medical revolution,” Per. Med., vol. 15, no. 5, pp. 429–448, 2018, doi: 10.2217/pme-2018-0044.

\bibitem{Yang2025} S.~Yang \textit{et al.}, “A deep learning approach to stress recognition through multimodal physiological signal image transformation,” \textit{Scientific Reports}, vol.~15, art.~no.~22258, 2025, doi:~10.1038/s41598-025-01228-3.

\bibitem{Gkikas2025MultiRep}
S.~Gkikas, I.~Kyprakis, and M.~Tsiknakis, “Multi-representation diagrams for pain recognition: Integrating various electrodermal activity signals into a single image,” in \textit{Companion Proc. 27th Int. Conf. on Multimodal Interaction (ICMI Companion)}, New York, NY, USA: ACM, 2025, pp.~162--171, doi:~10.1145/3747327.3764793.


\bibitem{Ziaratnia2024CCTLSTM}
S.~Ziaratnia, T.~Laohakangvalvit, M.~Sugaya, and P.~Sripian, “Multimodal deep learning for remote stress estimation using CCT-LSTM,” in \textit{Proc. IEEE/CVF Winter Conf. on Applications of Computer Vision (WACV)}, Waikoloa, HI, USA, 2024, pp.~8321--8329, doi:~10.1109/WACV57701.2024.00815.

\bibitem{Vos}
G.~Vos, K.~Trinh, Z.~Sarnyai, and M.~R.~Azghadi, ``Generalizable machine learning for stress monitoring from wearable devices: A systematic literature review,'' \emph{International Journal of Medical Informatics}, vol.~173, p.~105026, 2023, doi:~10.1016/j.ijmedinf.2023.105026.


\bibitem{Ilse2018MIL} M. Ilse, J. M. Tomczak, and M. Welling, “Attention-based deep multiple instance learning,” in Proceedings of the 35th International Conference on Machine Learning (ICML), Stockholm, Sweden, 2018, pp. 2127–2136.

\bibitem{Huang} H. Huang, N. Ardalani, A. Sun, L. Ke, H.-H. S. Lee, S. Bhosale, C.-J. Wu, and B. Lee, “Toward Efficient Inference for Mixture of Experts,” in Proc. 38th Conf. Neural Information Processing Systems (NeurIPS), 2024.

\bibitem{Fedus2022}
W.~Fedus, B.~Zoph, and N.~Shazeer, “Switch transformers: Scaling to trillion parameter models with simple and efficient sparsity,” \textit{J. Mach. Learn. Res.}, vol.~23, no.~1, art.~no.~120, pp.~1--39, Jan.~2022.


\bibitem{CARDIOCARE} “Home – CARDIOCARE.” [Online]. Available: \href{https://cardiocare-project.eu/}{https://cardiocare-project.eu/} Accessed: Dec. 2025.

\bibitem{Nechita2025}
L. C. Nechita, \emph{et al.}, “AI and smart devices in cardio-oncology: Advancements in cardiotoxicity prediction and cardiovascular monitoring,” \emph{Diagnostics}, vol. 15, no. 6, p. 787, 2025,
doi: 10.3390/diagnostics15060787.

\bibitem{Calvo}
A.~Calvo, J.~Martin, and C.~Martin, ``Early detection of chronic stress using wearable devices: A machine learning approach with the WESAD database,'' in \emph{Proc. 11th Int. Conf. Information and Communication Technologies for Ageing Well and e-Health (ICT4AWE)}, 2025, pp. 189–196, doi: 10.5220/0013209700003938.

\bibitem{Heyat2022}
M. B. Bin Heyat \emph{et al.}, “Wearable flexible electronics based cardiac electrode for researcher mental stress detection system using machine learning models on single lead electrocardiogram signal,” Biosensors, vol. 12, no. 6, p. 427, 2022, doi: 10.3390/bios12060427.

\bibitem{Amin2025}
O.~B.~Amin, V.~Mishra, T.~M.~Tapera, R.~Volpe, and A.~Sathyanarayana, ``Extending Stress Detection Reproducibility to Consumer Wearable Sensors,'' \emph{arXiv preprint arXiv:2505.05694}, 2025. [Online]. Available: https://arxiv.org/abs/2505.05694

\bibitem{Martinez2022}
G.~J. Martinez, T.~Grover, S.~M. Mattingly, \textit{et al.}, “Alignment between heart rate variability from fitness trackers and perceived stress: Perspectives from a large-scale in situ longitudinal study of information workers,” \textit{JMIR Human Factors}, vol.~9, no.~3, p.~e33754, Aug.~2022, doi:~10.2196/33754.

\bibitem{Arya2025}
P.~Arya, \emph{et al.}, ``Visualizing relaxation in wearables: Multi-domain feature fusion of HRV using fuzzy recurrence plots,'' \emph{Sensors}, vol.~25, no.~13, p.~4210, Jul.~2025, doi: 10.3390/s25134210.

\bibitem{GarminVenuSQ}
Garmin Ltd., ``Venu SQ smartwatch,'' [Online]. Available: https://www.garmin.com/en-US/p/707174/. Accessed: Dec. 2025.

\bibitem{PolarH10}
Polar Electro Oy, ``Polar H10 heart rate sensor,'' [Online]. Available: https://www.polar.com/en/sensors/h10-heart-rate-sensor. Accessed: Dec. 2025.

\bibitem{Cohen1983}
S.~Cohen, T.~Kamarck, and R.~Mermelstein, ``A global measure of perceived stress,'' \emph{Journal of Health and Social Behavior}, vol.~24, no.~4, pp.~385--396, 1983, doi: 10.2307/2136404.

\bibitem{neurokit}
D.~Makowski \emph{et al.}, ``NeuroKit2: A Python toolbox for neurophysiological signal processing,'' \emph{Behavior Research Methods}, vol.~53, no.~4, pp.~1689--1696, 2021, doi: 10.3758/s13428-020-01516-y.

\bibitem{Gemini} 
Google LLC, “Gemini,” AI image generation system. Accessed: Dec. 2025. [Online]. Available: https://gemini.google.com/

\bibitem{tiny}
S.~Gkikas, I.~Kyprakis, and M.~Tsiknakis, ``Tiny-BioMoE: A Lightweight Embedding Model for Biosignal Analysis,'' in \emph{Companion Proceedings of the 27th International Conference on Multimodal Interaction (ICMI Companion)}, New York, NY, USA: ACM, 2025, pp.~117--126, doi: 10.1145/3747327.3764788.

\end{thebibliography}
\end{document}